\documentclass[technicalnote]{IEEEtran}
\usepackage{graphicx}
\usepackage{algorithmic}
\usepackage{amsmath}
\usepackage{amsfonts}
\usepackage{array}
\newcommand{\PreserveBackslash}[1]{\let\temp=\\#1\let\\=\temp}
\newcolumntype{C}[1]{>{\PreserveBackslash\centering}p{#1}}
\newcolumntype{R}[1]{>{\PreserveBackslash\raggedleft}p{#1}}
\newcolumntype{L}[1]{>{\PreserveBackslash\raggedright}p{#1}}
\usepackage[usenames]{color}
\usepackage{tabularx}
\usepackage{multicol}
\usepackage{subfigure}
\usepackage{bm}
\usepackage{cite}
\usepackage[normalem]{ulem}

\begin{document}

\title{$l^1$-norm Penalized Orthogonal Forward Regression}
\author{Xia Hong, Sheng Chen, Yi Guo, and Junbin Gao %
\thanks{Xia Hong is with the School of Systems Engineering, University of Reading,
 Reading,  RG6 6AY,UK (E-mail: x.hong@reading.ac.uk).} %
 \thanks{Sheng Chen is with Electronics and Computer Science, University of Southampton,
 Southampton SO17 1BJ, UK (E-mail: sqc@ecssoton.ac.uk), and also with King Abdulaziz
 University, Jeddah 21589, Saudi Arabia.}%
\thanks{Yi Guo is with CSIRO Mathematics and Information Sciences, North Ryde, NSW 1670,
 Australia (E-mail: yg\_au@yahoo.com.au).}
\thanks{Junbin Gao is with the School of Computing and Mathematics, Charles Sturt
 University, Bathurst, NSW 2795, Australia (E-mail: jbgao@csu.edu.au).} %
\thanks{Junbin Gao and Xia Hong acknowledge the support of ARC under Grant DP130100364.}
\vspace*{-5mm}
}

 \maketitle

\begin{abstract}
 A $l^1$-norm penalized orthogonal forward regression ($l^1$-POFR) algorithm is
 proposed based on the concept of leave-one-out mean square error (LOOMSE). Firstly, a
 new $l^1$-norm penalized cost function is defined in the constructed orthogonal space,
 and each orthogonal basis is associated with an individually tunable regularization
 parameter. Secondly, due to orthogonal computation, the LOOMSE can be analytically
 computed without actually splitting the data set, and moreover a closed form of the
 optimal regularization parameter in terms of minimal LOOMSE is derived. Thirdly, a
 lower bound for regularization parameters is proposed, which can be used for robust
 LOOMSE estimation by adaptively detecting and removing regressors to an inactive set
 so that the computational cost of the algorithm is significantly reduced. Illustrative
 examples are included to demonstrate the effectiveness of this new $l^1$-POFR approach.
 \end{abstract}


\section{Introduction}\label{S1}

 One of the main aims in data modeling is good generalization, i.e. the model's capability
 to approximate accurately the system output for unseen data. Sparse models can be
 constructed using the $l^1$-penalized cost function, e.g. the basis pursuit or least
 absolute shrinkage and selection operator (LASSO) \cite{Chenss1998a,Tibshirani1996a,Efron2004a}.
 Based on a fixed single $l^1$-penalized regularization parameter, the LASSO can be configured
 as a standard quadratic programming optimization problem. By exploiting piecewise linearity
 of the problem, the least angle regression (LAR) procedure \cite{Efron2004a} is developed for
 solving the problem efficiently. Note that the computational efficiency in LASSO is facilitated
 by a \emph{single} regularization parameter setting. For more complicated constraints, e.g.
 multiple regularizers, the cross validation by actually splitting data sets as the means of
 evaluating model generalization comes with considerably large overall computational overheads.

 Alternatively the forward orthogonal least squares (OLS) algorithm efficiently constructs
 parsimonious models \cite{Chens1989a,Chen_etal1991}. Fundamental to the evaluation of model
 generalization capability is the concept of cross-validation \cite{Stone1974a}, and one
 commonly used version of cross-validation is the leave-one-out (LOO) cross validation. For
 the linear-in-the-parameters models, the LOO mean square error (LOOMSE) can be calculated
 without actually splitting the training data set and estimating the associated models, by
 making use of the Sherman-Morrison-Woodbury theorem. Using the LOOMSE as the model term
 selective criterion to seek the model generalization, an efficient orthogonal forward
 regression (OFR) procedure have been introduced \cite{Hong2003d}. Furthermore, the $l^2$-norm
 based regularization techniques \cite{Mackay1991a,Chen_etal1996,Orr1993a} have been
 incorporated into the OLS algorithm to produce a regularized OLS (ROLS) algorithm that
 carries out model term selection while reduces the variance of parameter estimate simultaneously
 \cite{Chens2003b}. The optimization of $l^1$-norm regularizer with respect to model
 generalization analytically is however less studied.

 In this contribution, we propose a $l^1$-norm penalized OFR ($l^1$-POFR) algorithm to
 carry out the regularizer optimization as well as model term selection and parameter
 estimation simultaneously in a forward regression manner. The algorithm is based on a new
 $l^1$-norm penalized cost function with multiple $l^1$ regularizers, each of which is
 associated with an orthogonal basis vector by orthogonal decomposition of the regression
 matrix of the selected model terms. We derive a closed form of the optimal regularization
 parameter in terms of minimal LOOMSE. To save computational costs an inactive set is used
 along the OFR process by predicting whether any model terms will be unselectable in future
 regression steps.

\section{Preliminaries}\label{S2}

 Consider the general nonlinear system represented by the nonlinear model \cite{Harris2002a,Chens1989b}:
\begin{equation}\label{eq:1}
 y(k) = f(\boldsymbol{x}(k)) + v(k) ,
\end{equation}
 where $\boldsymbol{x}(k)=\big[x_1(k) ~ x_2(k) \cdots x_m(k)\big]^{\rm T}\in \mathbb{R}^m$
 denotes the $m$-dimensional input vector at sample time index $k$ and $y(k)$ is the system
 output variable, respectively, while $v(k)$ denotes the system white noise and $f(\bullet )$
 is the unknown system mapping.

 The unknown system (\ref{eq:1}) is to be identified based on an observation data set
 $D_N=\{\boldsymbol{x}(k),y(k)\}_{k=1}^N$ using some suitable functional which can
 approximate $f(\bullet )$ with arbitrary accuracy. Without loss of generality, we use
 $D_N$ to construct a radial basis function (RBF) network model of the form
\begin{equation}\label{eq:2}
 \widehat{y}^{(M)}(k) = f^{(M)}(\boldsymbol{x}(k))
 = \sum\limits_{i=1}^{M}\theta_i \phi_i(\boldsymbol{x}(k)) ,
\end{equation}
 where $\widehat{y}^{(M)}(k)$ is the model prediction output for $\boldsymbol{x}(k)$
 based on the $M$-term RBF model, and $M$ is the total number of regressors or model terms,
 while $\theta_i$ are the model weights. The regressor $\phi_i(\boldsymbol{x})$ is given by
\begin{equation}\label{eq:3}
 \phi_i(\boldsymbol{x}) = \exp\Big(-\frac{\|\boldsymbol{x}- \boldsymbol{c}_i \|^2}{2\tau^2}
 \Big)
\end{equation}
 in which $\boldsymbol{c}_i=\big[c_{1,i} ~ c_{2,i} \cdots c_{m,i}\big]^{\rm T}$ is known
 as the center vector of the $i$th RBF unit and $\tau$ is an RBF width parameter. We assume
 that each RBF unit is placed on a training data, namely, all the RBF center vectors
 $\{\boldsymbol{c}_i\}_{i=1}^M$ are selected from the training data $\{\boldsymbol{x}(k)\}_{k=1}^N$,
 and the RBF width $\tau$ has been predetermined, for example, using cross validation.

 Let us denote $e^{(M)}(k)=y(k)-\widehat{y}^{(M)}(k)$ as the $M$-term modeling error for
 the input data $\boldsymbol{x}(k)$. Over the training data set $D_N$, further denote
 $\boldsymbol{y}=[y(1) ~ y(2) \cdots y(N)]^{\rm T}$, $\boldsymbol{e}^{(M)}=
 \big[e^{(M)}(1) ~ e^{(M)}(2) \cdots e^{(M)}(N)\big]^{\rm T}$, and $\boldsymbol{\Phi}_M
 =\big[\boldsymbol{\phi}_1 ~ \boldsymbol{\phi}_2 \cdots \boldsymbol{\phi}_{M}\big]$ with
 $\boldsymbol{\phi}_n=\big[\phi_n(\boldsymbol{x}(1)) ~ \phi_n(\boldsymbol{x}(2)) \cdots
 \phi_n(\boldsymbol{x}(N))]^{\rm T}$, $1\le n\le M$. We have the $M$-term model in the
 matrix form of
\begin{equation}\label{eq:4}
 \boldsymbol{y} = \boldsymbol{\Phi}_M \boldsymbol{\theta}_M + \boldsymbol{ e}^{(M)} ,
\end{equation}
 where $\boldsymbol{\theta}_M=\big[\theta_1 ~ \theta_2 \cdots \theta_M\big]^{\rm T}$.
 Let an orthogonal decomposition of the regression matrix $\boldsymbol{\Phi}_M$ be
\begin{equation}\label{eq:5}
 \boldsymbol{\Phi}_M = \boldsymbol{W}_M \boldsymbol{A}_M ,
\end{equation}
 where
\begin{equation}\label{eq:6}
 \boldsymbol{A}_M = \left[ \begin{array}{cccc} 1 & a_{1,2} & \cdots & a_{1,M} \\
  0 & 1 & \ddots & \vdots \\ \vdots & \ddots & \ddots &  a_{M-1,M} \\
  0 & \cdots & 0 & 1 \end{array} \right]
\end{equation}
 and
\begin{equation}\label{eq:7}
 \boldsymbol{W}_M=\big[\boldsymbol{w}_1 ~ \boldsymbol{w}_2 \cdots \boldsymbol{w}_M \big]
\end{equation}
 with columns satisfying $\boldsymbol{w}_i^{\rm T}\boldsymbol{w}_j=0$, if $i\ne j$.
 The regression model (\ref{eq:4}) can alternatively be expressed as
\begin{equation}\label{eq:8}
 \boldsymbol{y} = \boldsymbol{W}_M \boldsymbol{g}_M + \boldsymbol{e}^{(M)} ,
\end{equation}
 where the `orthogonal' model's weight vector $\boldsymbol{g}_M=\big[g_1 ~ g_2
 \cdots g_M\big]^{\rm T}$ satisfies the triangular system $\boldsymbol{A}_M
 \boldsymbol{\theta}_M=\boldsymbol{g}_M$, which can be used to determine the
 original model parameter vector $\boldsymbol{\theta}_M$, given $\boldsymbol{A}_M$
 and $\boldsymbol{g}_M$.

 Further consider the following weighted $l^1$-norm penalized OLS criterion for
 the model (\ref{eq:8})
\begin{equation}\label{eq:9}
 L_e\big(\boldsymbol{\Lambda}_M,\boldsymbol{g}_M\big) =
 \big\| \boldsymbol{y} - \boldsymbol{W}_M\boldsymbol{g}_M \big\|^2
 + \sum_{i=1}^{M}\lambda_i \big| g_i\big| ,
\end{equation}
 where $\boldsymbol{\Lambda}_M=\mbox{diag}\{\lambda_1,\lambda_2,\cdots ,\lambda_M\}$,
 which contains the local regularization parameters $\lambda_i\ge \varepsilon$, for
 $1\le i\le M$, and $\varepsilon >0$ is a predetermined lower bound for the regularization
 parameters. For a given $\boldsymbol{\Lambda}_M$, the solution for $\boldsymbol{g}_M$
 can be obtained by setting the subderivative vector of $L_e$ to zero, i.e.
 ${\partial L_e \over \partial \boldsymbol{g}_M}= \boldsymbol{0}$, yielding
\begin{equation}\label{eq:10}
 g_i ^{({\rm olasso})} = \left( \big|g_i^{({\rm LS})}\big| - \frac{\lambda_i/2}{\boldsymbol{w}^{\rm T}_i
 \boldsymbol{w}_i}\right)_{+} \mbox{sign} \big(g_i^{({\rm LS})}\big)
\end{equation}
 for $1\le i\le M$, with the usual least squares solution given by $g_i^{({\rm LS)}}=
 \frac{\boldsymbol{w}^{\rm T}_i\boldsymbol{y}}{\boldsymbol{w}^{\rm T}_i\boldsymbol{w}_i}$,
 and the operator $( ~ )_{+}$
\begin{equation}\label{eq:11}
 z_+=\left\{\begin{array}{cc}
  z , & \mbox{if } z > 0 , \\
  0 , & \mbox{if } z \le 0 .
 \end{array}\right.
\end{equation}

 Unlike the LASSO \cite{Chenss1998a,Tibshirani1996a}, our objective
 $L_e\big(\boldsymbol{\Lambda}_M,\boldsymbol{g}_M\big)$ is constructed on the
 orthogonal space and the $l^1$-norm parameter constraints are associated with
 the orthogonal bases $\boldsymbol{w}_i$, $1\le i\le M$. Since the cost function
 (\ref{eq:9}) contains sparsity inducing $l^1$ norm, some parameters
 $g_i^{({\rm olasso})}$ will be returned as zeros, producing a sparse model in
 the orthogonal space spanned by the columns of $\boldsymbol{W}_M$, which
 corresponds to a sparse model in the original space spanned by the columns of
 $\boldsymbol{\Phi}_M$.

\section{Regularization parameter optimization and model construction with LOOMSE}\label{S3}

 Each OFR stage involves the joint regularization parameter optimization, model
 term selection and parameter estimation. The regularization parameters with respect
 to their associated candidate regressors are optimized using the approximate LOOMSE
 formula that is derived in Section~\ref{S3:2}, and  the regressor with the smallest
 LOOMSE is selected.

\subsection{Model representation and LOOMSE in $n$-th stage OFR}\label{S3:1}

 Consider the OFR modeling process that has produced the $n$-term model. Let us denote
 the constructed $n$ columns of regressors as $\boldsymbol{W}_n=\big[\boldsymbol{w}_1 ~
 \boldsymbol{w}_2 \cdots \boldsymbol{w}_n\big]$, with $\boldsymbol{w}_n=\big[w_n(1) ~
 w_n(2) \cdots w_n(N)\big]^{\rm T}$. The model output vector of this $n$-term model is
 given by
\begin{equation}\label{eq:12}
 \boldsymbol{\widehat{y}}^{(n)} = \sum_{i=1}^n g_i^{({\rm olasso})} \boldsymbol{w}_i ,
\end{equation}
 and the corresponding modeling error vector by $\boldsymbol{e}^{(n)}=
 \boldsymbol{y}-\boldsymbol{\widehat{y}}^{(n)}$. Clearly, the $n$th OFR stage can
 be represented by
\begin{equation}\label{eq:13}
 \boldsymbol{e}^{(n-1)} = g_n\boldsymbol{w}_n + \boldsymbol{e}^{(n)} .
\end{equation}
 The model form (\ref{eq:13}) illustrates the fact that the $n$th OFR stage is simply
 to fit a one-variable model using the current model residual produced after the
 $(n-1)$th stage as the desired system output. Since  $\boldsymbol{w}^{\rm T}_n
 \boldsymbol{{\hat y}}^{(n-1)}=0$, it is easy to verify that $g_n^{({\rm LS})}=
 \frac{\boldsymbol{w}^{\rm T}_n\boldsymbol{y}}{\boldsymbol{w}^{\rm T}_n\boldsymbol{w}_n}
 =\frac{\boldsymbol{w}^{\rm T}_n\boldsymbol{e}^{(n-1)}}{\boldsymbol{w}^{\rm T}_n\boldsymbol{w}_n}$.

 The selection of one regressor from the candidate regressors involves initially
 generating candidate $\boldsymbol{w}_n$ by making each candidate regressor to be
 orthogonal to the $(n-1)$ orthogonal basis vectors, $\boldsymbol{w}_i$ for $1\le i\le n-1$
 obtained in the previous $(n-1)$ OFR stages, followed by evaluating their contributions.
 Consider the case of $2 \big| \boldsymbol{w}_n^{\rm T}\boldsymbol{e}^{(n-1)}\big| >
 \varepsilon$. Applying (\ref{eq:10}) to (\ref{eq:13}), we note that clearly as
 $\lambda_n$ decreases away from $2\big|\boldsymbol{w}_n^{\rm T}\boldsymbol{e}^{(n-1)}\big|$
 towards $\varepsilon$, $g_n^{({\rm olasso})}$ increases its magnitude at a linear
 rate to $\lambda_n$, from zero to an upper bound $\big|g_n^{({\rm B})}\big|$ with
\begin{equation}\label{eq:14}
 g_n^{({\rm B})}=\Big( \big|g_n^{({\rm LS})}\big| - \frac{\varepsilon}
 {2\boldsymbol{w}_n^{\rm T}\boldsymbol{w}_n}\Big)_{+} \mbox{sign}\big(g_n^{({\rm LS})}\big) .
\end{equation}
 For any candidate regressor, it is vital that we evaluate its potential model
 generalization performance using the most suitable value of $\lambda_n$. The
 optimization of the LOOMSE with respect to $\lambda_n$ is detailed in
 Section~\ref{S3:2}, based on the idea of the LOO cross validation outlined below.

 Suppose that we sequentially set aside each data point in the estimation set $D_N$ in
 turn and estimate a model using the remaining $(N-1)$ data points. The prediction error
 is calculated on the data point that has not been used in estimation. That is, for
 $k=1,2,\cdots, N$, the models are estimated based on $D_N\setminus (\boldsymbol{x}(k),y(k))$,
 respectively, and the outputs are denoted as $\widehat{y}^{(n-1,-k)}(k, \lambda_n)$. Then,
 the LOO prediction error based on the $k$th data sample is calculated as
\begin{equation}\label{eq:15}
 e^{(n,-k)}(k,\lambda_n)= y(k) -\widehat{y}^{(n-1,-k)}(k, \lambda_n).
\end{equation}
 The LOOMSE is defined as the average of all these prediction errors, given by
 $J\big(\lambda_n\big) =E\left[ \big(e^{(n,-k)}(k, \lambda_n)\big)^2 \right]$. Thus the
 optimal regularization parameter for the $n$th stage is given by
\begin{equation}\label{eq:16}
 \lambda_n^{\rm opt}=\arg \, \min_{\lambda_n}\Big\{J\big(\lambda_n\big) =
 \frac{1}{N} \sum_{k=1}^{N}\big( e^{(n,-k)}(k, \lambda_n)\big)^2 \Big\} .
\end{equation}
 Evaluation of $J\big(\lambda_n\big)$ by directly splitting the data set requires
 extensive computational efforts. Instead, we show in Section~\ref{S3:2} that
 $J\big(\lambda_n\big)$ can be approximately calculated without actually
 sequentially splitting the estimation data set. Furthermore, we also show that
 the optimal value $\lambda_n^{\rm opt}$ can be obtained in a closed-form expression.

\subsection{Optimal regularization parameter estimate}\label{S3:2}

 We notice from (\ref{eq:10}) that $g_n^{({\rm olasso})}=0$ if $2\big|\boldsymbol{w}_n^{\rm T}
 \boldsymbol{e}^{(n-1)}\big| < \lambda_n$, and thus a sufficient condition that a given
 $\boldsymbol{w}_n$ may be excluded from the candidate pool without explicitly determining
 $\lambda_n$ is $2\big|\boldsymbol{w}_n^{\rm T} \boldsymbol{e}^{(n-1)}\big| < \varepsilon$,
 which is the regularizer's lower bound, a preset value indicating the correlation of the
 candidate regressor. Hence, in the following we assume that $2\big|\boldsymbol{w}_n^{\rm T}
 \boldsymbol{e}^{(n-1)}\big| > \varepsilon$, and we have
\begin{equation}\label{eq:17}
 \boldsymbol{g}_n^{({\rm olasso})} = \boldsymbol{H}_n^{-1} \left(\boldsymbol{W}_n^{\rm T}
 \boldsymbol{y} - \boldsymbol{\Lambda}_n\mbox{sign}(\boldsymbol{g}_n^{({\rm LS})})/2 \right) ,
\end{equation}
 where $\boldsymbol{g}_n^{({\rm olasso})}=\big[g_1^{({\rm olasso})} ~ g_2^{({\rm olasso})}
 \cdots g_n^{({\rm olasso})}\big]^{\rm T}$, $\mbox{sign}(\boldsymbol{g}_n)$ $=\big[\mbox{sign}(g_1)
 ~ \mbox{sign}(g_2)\cdots \mbox{sign}(g_n)\big]^{\rm T}$, and $\boldsymbol{H}_n=
 \boldsymbol{W}_n^{{\rm T}}\boldsymbol{W}_n$. Note that (\ref{eq:17}) is consistent to
 (\ref{eq:10}) for all terms with nonzero $g_i$. In the OFR procedure, any candidate terms
 $\boldsymbol{w}_i$ producing zero $g_i^{({\rm olasso})}$ will not be selected since they
 will not contribute to any reduction in the LOOMSE.

 The model residual is defined by
\begin{align}\label{eq:18}
 & \hspace*{-2mm}e^{(n)}(k,\lambda_n) = y(k) - \big(\boldsymbol{g}^{({\rm olasso})}\big)^{\rm T}
 \boldsymbol{w}(k) \nonumber \\ &= y(k) -
 \Big(\boldsymbol{y}^{\rm T}\boldsymbol{W}_n - \big(\mbox{sign}
 \big(\boldsymbol{g}^{({\rm LS})}\big)\big)^{\rm T}\boldsymbol{\Lambda}_n/2\Big)
 \boldsymbol{H}_n^{-1}\boldsymbol{w}(k) ,
\end{align}
 where $\boldsymbol{w}(k)$ denotes the transpose of the $k$th row of $\boldsymbol{W}_n$.
 If the data sample indexed at $k$ is removed from the estimation data set, the LOO
 parameter estimator obtained by using only the $(N-1)$ remaining data points is given by
\begin{align}\label{eq:19}
 \boldsymbol{ g}_n^{({\rm olasso},-k)} =& \big(\boldsymbol{H}_n^{(-k)}\big)^{-1}
 \Big(\big(\boldsymbol{W}_n^{(-k)}\big)^{\rm T}\boldsymbol{y}^{(-k)} - \nonumber \\ &
 \boldsymbol{\Lambda}_n\mbox{sign}\big(\boldsymbol{g}^{({\rm LS},-k)}\big)/2 \Big)
\end{align}
 in which $\boldsymbol{H}_n^{(-k)}=\big(\boldsymbol{W}_n^{(-k)}\big)^{\rm T}
 \boldsymbol{W}^{(-k)}_n$, $\boldsymbol{W}_n^{(-k)}$ and $\boldsymbol{y}^{(-k)}$
 denote the resultant regression matrix and desired output vector, respectively. It
 follows that we have
\begin{equation}\label{eq:21}
 \boldsymbol{ H}_n^{(-k)} = \boldsymbol{H}_n - \boldsymbol{w}(k)\boldsymbol{w}^{\rm T}(k) ,
\end{equation}
\begin{equation}\label{eq:22}
 \big(\boldsymbol{y}^{(-k)}\big)^{\rm T}\boldsymbol{W}_n^{(-k)} = \boldsymbol{y}^{\rm T}
 \boldsymbol{W}_n - y(k)\boldsymbol{w}^{\rm T}(k) .
\end{equation}
 The LOO error evaluated at $k$ is given by
\begin{align}\label{eq:20}
 e^{(n,-k)}(k,\lambda_n) &= y(k) - \big(\boldsymbol{g}^{({\rm olasso},-k)}\big)^{\rm T}
 \boldsymbol{w}(k) \nonumber \\
 & \hspace*{-10mm}= y(k) - \Big(\big(\boldsymbol{y}^{(-k)}\big)^{\rm T}\boldsymbol{W}_n^{(-k)} - \nonumber \\ &
 \hspace*{-6mm}\big(\mbox{sign}\big(\boldsymbol{g}^{({\rm LS},-k)}\big)\big)^{\rm T}
 \boldsymbol{\Lambda}_n/2 \Big) \big(\boldsymbol{H}_n^{(-k)}\big)^{-1}\boldsymbol{w}(k).
\end{align}

 Applying the matrix inversion lemma to (\ref{eq:21}) yields
\begin{align}\label{eq:23}
 \big(\boldsymbol{H}_n^{(-k)}\big)^{-1} =& \big(\boldsymbol{H}_n - \boldsymbol{w}(k)
 \boldsymbol{w}^{\rm T}(k)\big)^{-1} \nonumber \\ =&
 \boldsymbol{H}_n^{-1} + \frac{\boldsymbol{H}_n^{-1}\boldsymbol{w}(k)\boldsymbol{w}^{\rm T}(k)
 \boldsymbol{H}_n^{-1}}{1 - \boldsymbol{w}^{\rm T}(k)\boldsymbol{H}_n^{-1}\boldsymbol{w}(k)}
\end{align}
 and
\begin{equation}\label{eq:24}
 \big(\boldsymbol{H}_n^{(-k)}\big)^{-1}\boldsymbol{w}(k) = \frac{\boldsymbol{H}_n^{-1}
 \boldsymbol{w}(k)}{1 - \boldsymbol{w}^{\rm T}(k)\boldsymbol{H}_n^{-1}\boldsymbol{w}(k)} .
\end{equation}
 Substituting (\ref{eq:22}) and (\ref{eq:24}) into (\ref{eq:20}) yields
\begin{align}\label{eq:25}
 & \hspace*{-2mm}e^{(n,-k)}(k,\lambda_n) = y(k) - \Big(\boldsymbol{y}^{\rm T}\boldsymbol{W}_n - y(k)
 \boldsymbol{w}^{\rm T}(k) - \nonumber \\ &
 \hspace*{10mm}\big(\mbox{sign}\big(\boldsymbol{g}^{({\rm LS },-k)}\big)\big)^{\rm T}
 \boldsymbol{\Lambda}_n/2\Big)\frac{\boldsymbol{H}_n^{-1}\boldsymbol{w}(k)}{1 -
 \boldsymbol{w}^{\rm T}(k)\boldsymbol{H}_n^{-1}\boldsymbol{w}(k)} \nonumber \\ & =
 \frac{y(k) - \Big(\boldsymbol{y}^{\rm T}\boldsymbol{W}_n - \big(
 \mbox{sign}\big(\boldsymbol{g}^{{\rm LS},-k)}\big)\big)^{\rm T}\boldsymbol{\Lambda}_n/2\Big)
 \boldsymbol{H}_n^{-1}\boldsymbol{w}(k)}{1 - \boldsymbol{w}^{\rm T}(k)\boldsymbol{H}_n^{-1}
 \boldsymbol{w}(k)} .
\end{align}
 Assuming that $\mbox{sign}\big(\boldsymbol{g}_n^{({\rm LS},-k)}\big)=
 \mbox{sign}\big(\boldsymbol{g}_n^{({\rm LS})}\big) $ holds for most data samples
 and then applying (\ref{eq:18}) to (\ref{eq:25}), we have
\begin{equation}\label{eq:26}
 e^{(n,-k)}(k,\lambda_n) = \gamma_n(k) e^{(n)}(k,\lambda_n) ,
\end{equation}
 where $\gamma_n(k)=\frac{1}{1-\sum_{i=1}^{n}\big(w_i(k)\big)^2\big/
 \boldsymbol{w}_i^{\rm T}\boldsymbol{w}_i}>0$, and $w_i(k)$ is the $k$th element
 of $\boldsymbol{w}_i$. The LOOMSE can then be calculated as
\begin{align}\label{eq:27}
 J\big(\lambda_n\big) =& \frac{1}{N}\sum_{k=1}^N \gamma_n^2(k) \big(e^{(n)}(k,\lambda_n)\big)^2 .
\end{align}
 We point out that in order for $\mbox{sign}\big(\boldsymbol{g}_n^{({\rm LS},-k)}\big)$
 and $\mbox{sign}\big(\boldsymbol{g}_n^{({\rm LS})}\big)$ to be different, each element
 in $\boldsymbol{g}_n^{({\rm LS})}$ needs to be very close to zero, which is unlikely
 since only the model terms satisfying $\big|\boldsymbol{w}_n^{\rm T}
 \boldsymbol{e}^{(n-1)}\big| >\varepsilon /2$ are considered. Hence we can treat
 $J\big(\lambda_n\big)$ given in (\ref{eq:27}) as the exact LOOMSE for any preset
 $\varepsilon$ that is not too small.

 We further represent (\ref{eq:18}) as
\begin{equation}\label{eq:28}
 e^{(n)}(k,\lambda_n) = \eta(k) + \frac{\lambda_n}{2\boldsymbol{w}_n^{\rm T}\boldsymbol{w}_n}
 w_n(k) \mbox{sign}\big(g_n^{({\rm LS})}\big) ,
\end{equation}
 where $\eta(k)=e^{(n-1)}(k)- g_n^{({\rm LS})}w_n(k)$ is the model residual
 obtained based on the least square estimate at the $n$th step stage. By setting
 ${\partial J(\lambda_n)\over \partial\lambda_n}=0$, we obtain $\lambda_n$ in the
 form of the weighted least square estimate
\begin{equation}\label{eq:29}
 \lambda_n = -2\mbox{sign}\big(g_n^{({\rm LS})}\big) \boldsymbol{w}_n^{\rm T}
 \boldsymbol{w}_n \boldsymbol{w}_n^{\rm T} \boldsymbol{\Gamma}^{(n)}
 \boldsymbol{\eta} \big/ \boldsymbol{w}_n^{\rm T} \boldsymbol{\Gamma}^{(n)}\boldsymbol{w}_n ,
\end{equation}
 where $\boldsymbol{\Gamma}^{(n)}=\mbox{diag}\big\{\gamma_n^2(1),\gamma_n^2(2),
 \cdots,\gamma_n^2(N)\big\}$ and $\boldsymbol{\eta}=\big[\eta (1) ~ \eta (2)
 \cdots \eta (N)\big]^{\rm T}\in \mathbb{R}^N$. Finally we calculate
\begin{align}\label{eq:30}
 \lambda_n^{\rm opt} =& \max \Big\{ \min \Big\{ 2\big| \boldsymbol{w}_n^{\rm T}
 \boldsymbol{e}^{(n-1)}\big|,
 -2\mbox{sign}\big(g_n^{({\rm LS})}\big)\boldsymbol{w}_n^{\rm T}\boldsymbol{w}_n
 \nonumber \\ & \times
 \boldsymbol{w}_n^{\rm T}\boldsymbol{\Gamma}^{(n)}\boldsymbol{\eta}\big/
 \boldsymbol{w}_n^{\rm T}\boldsymbol{\Gamma}^{(n)}\boldsymbol{w}_n\Big\} ,
 \varepsilon \Big\} ,
\end{align}
 in order to satisfy the constraint that $\varepsilon \le \lambda_n^{\rm opt}
 \le 2 \big| \boldsymbol{w}_n^{\rm T}\boldsymbol{e}^{(n-1)}\big|$. For
 $\lambda_n^{\rm opt}$ obtained using (\ref{eq:30}), we consider the following
 two cases:
\begin{enumerate}
\item If $\lambda_n^{\rm opt}=2\big| \boldsymbol{w}_n^{\rm T}\boldsymbol{e}^{(n-1)}\big|$,
 then $g_n^{({\rm olasso})}=0$, and this candidate regressor will not be selected.
\item If $\varepsilon \le \lambda_n^{\rm opt} <2\big| \boldsymbol{w}_n^{\rm T}
 \boldsymbol{e}^{(n-1)}\big|$, then calculate $J\big(\lambda_n^{\rm opt}\big)$
 based on $(\ref{eq:27})$ as the LOOMSE for this candidate regressor.
\end{enumerate}

\subsection{Moving unselectable regressors to the inactive set}\label{S3:3}

 From Section~\ref{S3:2} we noted that a candidate regressor satisfying
 $2\big|\boldsymbol{w}_n^{\rm T}\boldsymbol{e}^{(n-1)}\big| < \varepsilon$ does
 not need to be considered at the $n$th stage of selection. To save computational
 cost, we define the inactive set ${\mathcal S}$ as the index set of the unselectable
 regressors removed from the pool of candidates.

 In the $n$th OFR stage, all the candidate regressors in the candidate pool are made
 orthogonal to the previously selected $(n-1)$ regressors, and the candidate with the
 smallest LOOMSE value is selected as the $n$th model term $\boldsymbol{w}_n$. Denote
 any other candidate regressor as $\boldsymbol{w}^{(-)}$.

\emph{Main Results}:
 If $\big\|\boldsymbol{w}^{(-)}\big\|\cdot \big\|\boldsymbol{e}^{(n-1)}\big\| <
 \frac{\varepsilon}{2}$, then this candidate regressor will never be selected in further
 regression stages, and hence it can be moved to ${\mathcal S}$.

\emph{Proof}: At the $(n+1)$th OFR stage, consider making the regressor
 $\boldsymbol{w}^{(-)}$ orthogonal to $\boldsymbol{w}_n$, and define
\begin{equation}\label{eq:31}
 \boldsymbol{w}^{(+)} = \boldsymbol{w}^{(-)} - \frac{\boldsymbol{w}_n^{\rm T}
 \boldsymbol{w}^{(-)}}{\boldsymbol{w}_n^{\rm T}\boldsymbol{w}_n} \boldsymbol{w}_n .
\end{equation}
 Clearly,
\begin{align}\label{eq:32}
 \big\| \boldsymbol{w}^{(+)}\big\|^2 =& \Big(\boldsymbol{w}^{(-)} -
 \frac{\boldsymbol{w}_n^{\rm T}\boldsymbol{w}^{(-)}}{\boldsymbol{w}_n^{\rm T}\boldsymbol{w}_n}
 \boldsymbol{w}_n\Big)^{\rm T} \Big(\boldsymbol{w}^{(-)} - \frac{\boldsymbol{w}_n^{\rm T}
 \boldsymbol{w}^{(-)}}{\boldsymbol{w}_n^{\rm T}\boldsymbol{w}_n} \boldsymbol{w}_n\Big) \nonumber \\
 =& \big\|\boldsymbol{w}^{(-)}\big\|^2 - \frac{\big(\boldsymbol{w}_n^{\rm T}
 \boldsymbol{w}^{(-)}\big)^2}{\boldsymbol{w}_n^{\rm T}\boldsymbol{w}_n} \le
 \big\|\boldsymbol{w}^{(-)}\big\|^2 .
\end{align}
 The model residual vector after the selection of $\boldsymbol{w}_n$ is
\begin{equation}\label{eq:33}
 \boldsymbol{e}^{(n)}=\boldsymbol{e}^{(n-1)} - g_n^{({\rm olasso})}\boldsymbol{w}_n ,
\end{equation}
 where $g_n^{({\rm olasso})}$ can be written as
\begin{equation}\label{eq:34}
 g_n^{({\rm olasso})} = \Big(\boldsymbol{w}_n^{\rm T}\boldsymbol{e}^{(n-1)} -
 \frac{\lambda_n}{2}\mbox{sign}\big(g_n^{({\rm LS})}\big) \Big) \big/
 \boldsymbol{w}_n^{\rm T}\boldsymbol{w}_n .
\end{equation}
 Thus we have
\begin{align}\label{eq:35}
 \big\|\boldsymbol{e}^{(n)}\big\|^2 =& \big\|\boldsymbol{e}^{(n-1)}\big\|^2
 - 2 g_n^{({\rm olasso})}\boldsymbol{w}_n^{\rm T}\boldsymbol{e}^{(n-1)} \nonumber \\ &
 + \big(g_n^{({\rm olasso})}\big)^2\boldsymbol{w}_n^{\rm T}\boldsymbol{w}_n ,
\end{align}
\begin{align}\label{eq:36}
 \big(g_n^{({\rm olasso})}\big)^2\boldsymbol{w}_n^{\rm T}\boldsymbol{w}_n
 =& \Big(\big(\boldsymbol{w}_n^{\rm T}\boldsymbol{e}^{(n-1)}\big)^2 - \nonumber \\ &
 \hspace*{-10mm}\lambda_n \mbox{sign}\big(g_n^{{\rm LS})}\big) \boldsymbol{w}_n^{\rm T}
 \boldsymbol{e}^{(n-1)} + \frac{\lambda_n^2}{4}\Big)\big/ \boldsymbol{w}_n^{\rm T}
 \boldsymbol{w}_n ,
\end{align}
 and
\begin{align}\label{eq:37}
 2 g_n^{({\rm olasso})} \boldsymbol{w}_n^{\rm T}\boldsymbol{e}^{(n-1)} =&
 \Big(2\big(\boldsymbol{w}_n^{\rm T}\boldsymbol{e}^{(n-1)}\big)^2 - \nonumber \\ &
 \hspace*{-10mm}\lambda_n \mbox{sign}\big(g_n^{({\rm LS})}\big) \boldsymbol{w}_n^{\rm T}
 \boldsymbol{e}^{(n-1)} \Big)\big/ \boldsymbol{w}_n^{\rm T}\boldsymbol{w}_n .
\end{align}
 Substituting (\ref{eq:36}) and (\ref{eq:37}) into (\ref{eq:35}) yields
\begin{align}\label{eq:38}
 \big\|\boldsymbol{e}^{(n)}\big\|^2 =& \big\|\boldsymbol{e}^{(n-1)}\big\|^2 -
 \Big( \big(\boldsymbol{w}_n^{\rm T}\boldsymbol{e}^{(n-1)}\big)^2 -
 \frac{\lambda_n^2}{4} \Big) \big/ \boldsymbol{w}_n^{\rm T}\boldsymbol{w}_n \nonumber \\
 < & \big\|\boldsymbol{e}^{(n-1)}\big\|^2 ,
\end{align}
 due to the fact that $\big|\boldsymbol{w}_n^{\rm T}\boldsymbol{e}^{(n-1)}\big|
 > \frac{\lambda_n}{2}$. From (\ref{eq:32}) and (\ref{eq:38}), it can be
 concluded that
\begin{equation}\label{eq:39}
 \big\|\boldsymbol{w}^{(+)}\big\|\cdot\big\|\boldsymbol{e}^{(n)}\big\| <
 \big\|\boldsymbol{w}^{(-)}\big\|\cdot\big\|\boldsymbol{e}^{(n-1)}\big\| <
 \frac{\varepsilon}{2} .
\end{equation}
 Since $\big\|\boldsymbol{w}^{(+)}\big\|\cdot\big\|\boldsymbol{e}^{(n)}\big\|$
 is the upper bound of $\Big|\big(\boldsymbol{w}^{(+)}\big)^{\rm T}
 \boldsymbol{e}^{(n)}\Big|$, this means that this regressor will not be
 selected at the $(n+1)$th stage. By induction, it will never be selected
 in further regression stages, and hence it can be moved to ${\mathcal S}$.

\section{The proposed $l^1$-POFR algorithm}\label{S4}

 The proposed $l^1$-POFR algorithm integrates (i)~the model regressor selection based
 on minimizing the LOOMSE; (ii)~regularization parameter optimization also based on
 minimizing the LOOMSE; and (iii)~the mechanism of removing unproductive candidate
 regressors during the OFR procedure. Define
\begin{equation}\label{eq:40}
 \boldsymbol{\Phi}^{(n-1)} = \big[ \boldsymbol{w}_1 \cdots \boldsymbol{w}_{n-1} ~
 \boldsymbol{\phi}_n^{(n-1)} \cdots \boldsymbol{\phi}_M^{(n-1)} \big]\in \mathbb{R}^{N \times M} ,
\end{equation}
 with $\boldsymbol{\Phi }^{(0)}=\boldsymbol{\Phi}_M$. If some of the columns in
 $\boldsymbol{\Phi}^{(n-1)}$ have been interchanged, this will still be referred
 as $\boldsymbol{\Phi }^{(n-1)}$ for notational simplicity.
\begin{table}[hp!]
\vspace{-1mm}
\caption{The $n$th stage of the selection procedure.}
\label{tab:1}
\begin{center}
\vspace{-8mm}
\begin{tabular} {|p{8.5cm}|}\hline
 For $\{n \le j \le M\}
 \cap \{j \notin {\mathcal S}\}$,  denote the $k$th element of $\boldsymbol{\phi}_{j}^{(n-1)}$
 as $\phi_{j}^{(n-1)}(k)$ and compute $\alpha_j=\big(\boldsymbol{\phi}_{j}^{(n-1)}\big)^{\rm T}
 \boldsymbol{e}^{(n-1)}$, and $\beta_j=\big\|\boldsymbol{\phi}_{j}^{(n-1)}\big\|\cdot
 \big\|\boldsymbol{e}^{(n-1)}\big\|$.

 Step~1):~If $\beta_j < \varepsilon /2$, ${\mathcal S}={\mathcal S} \cup j $;
 Else if $\big|\alpha_j\big|< \varepsilon /2$, set $J_n^{(j)}$ as a very large
 positive number so that it will not be selected in Step~4).  Otherwise goto step~2).

 Step~2):~Calculate
\begin{eqnarray}
 \kappa_n^{(j)} &=& \big(\boldsymbol{\phi}_{j}^{(n-1)}\big)^{\rm T}
  \boldsymbol{\phi}_{j}^{(n-1)} , \label{eq:41} \\
 g_n^{({\rm LS},j)} &=& \frac{\alpha_j}{\kappa_n^{(j)}} , \label{eq:42} \\
 \boldsymbol{\Gamma}^{(n,j)} &=& \mbox{diag}\left\{\frac{1}{\Big(\zeta^{(n-1)}(1)- \big(\phi_{j}^{(n-1)}(1)\big)^2\big/
  \kappa_n^{(j)}\Big)^2} , \right. \nonumber \\ & & \hspace*{-10mm}
  \frac{1}{\Big(\zeta^{(n-1)}(2)- \big(\phi_{j}^{(n-1)}(2)\big)^2\big/
  \kappa_n^{(j)}\Big)^2} , \cdots , \nonumber \\ & & \hspace*{-10mm}
 \left. \frac{1}{\Big(\zeta^{(n-1)}(N)-\big(\phi_{j}^{(n-1)}(N)\big)^2\big/
  \kappa_n^{(j)}\Big)^2} \right\} \in \mathbb{R}^{N \times N} , \label{eq:43} \\
 \boldsymbol{\eta}^{(j)} &=& \boldsymbol{e}^{(n-1)} - g_n^{({\rm LS},j)}
  \boldsymbol{\phi}_{j}^{(n-1)} , \label{eq:44} \\
 \lambda_n^{({\rm opt},j)} & =& \max\Big\{ \min\Big\{2\big|\alpha_j\big|, -2
  \mbox{sign}\big(g_n^{({\rm LS},j)}\big) \kappa_n^{(j)} \nonumber \\ & &
  \hspace*{-12mm}\big(\boldsymbol{\phi}_{j}^{(n-1)}\big)^{\rm T}
  \boldsymbol{\Gamma}^{(n,j)} \boldsymbol{\eta}^{(j)} \big/\big(\boldsymbol{\phi}_{j}^{(n-1)}\big)^{\rm T}
  \boldsymbol{\Gamma}^{(n,j)}\boldsymbol{\phi}_{j}^{(n-1)}\Big\},\varepsilon\Big\} . \label{eq:45}
\end{eqnarray}

 Step~3):~If $\lambda_n^{({\rm opt},j)}=2\big|\alpha_j\big|$, set $J_n^{(j)}$
 as a very large positive number so that it will not be selected in Step~4);
 Otherwise calculate
\begin{eqnarray}
 g_n^{({\rm olasso},j)} &=& \big( \big|g_n^{({\rm LS},j)}\big| -
  \frac{\lambda_n^{({\rm opt},j)}/2}{\kappa_n^{(j)} }\big)_{+}
  \mbox{sign}\big(g_n^{({\rm LS},j)}\big) , \label{eq:46}
\end{eqnarray}
\begin{eqnarray}
 \boldsymbol{e}^{(n,j)} &=& \boldsymbol{e}^{(n-1)} - g_n^{({\rm olasso},j)}
  \boldsymbol{\phi}_{j}^{(n-1)} , \label{eq:47} \\
 J_n^{(j)} &=& \big(\boldsymbol{e}^{(n,j)}\big)^{\rm T}\boldsymbol{\Gamma}^{(n,j)}
  \boldsymbol{e}^{(n,j)}/N . \label{eq:48}
\end{eqnarray}

 Step~4):~Find
\begin{equation}\label{eq:49}
 J_n = J_n^{(j_n)} = \min \left\{ J_n^{(j)} , \ \{l \le j \le M\} \cap
  \{j \notin {\mathcal S}\} \right\} .
\end{equation}
 Then update $\boldsymbol{e}^{(n)}$ and $g_n^{({\rm olasso})}$ as
 $\boldsymbol{e}^{(n,j_n)}$ and $g_n^{({\rm olasso},j_n)}$, respectively. The
 $j_n$th and the $n$th columns of $\boldsymbol{\Phi }^{(n-1)}$ are interchanged,
 while the $j_n$th  column and the $n$th column of $\boldsymbol{A}_M$ are
 interchanged up to the $(n-1)$th row. This effectively selects the $n$th
 regressor in the subset model. The modified Gram-Schmidt orthogonalisation
 procedure \cite{Chens1989a} then calculates the $n$th row of the matrix
 $\boldsymbol{A}_M$ and transfers $\boldsymbol{\Phi }^{(n-1)}$ into
 $\boldsymbol{\Phi }^{(n)}$ as follows
\begin{equation}\label{eq:50}
 \!\!\! \left. \!\!\! \begin{array}{l}
 \boldsymbol{w}_n = \boldsymbol{\phi}_{n}^{(n-1)} ,  \\
 a_{n,j} = \boldsymbol{w}_n^{\rm T}\boldsymbol{\phi}_j^{(n-1)} \big/
 \boldsymbol{w}_n^{\rm T}\boldsymbol{w}_n , \{n+1  \le j \le M \}\cap
 \{j \notin {\mathcal S}\} , \\
 \boldsymbol{\phi}_j^{(n)} = \boldsymbol{\phi}_j^{(n-1)} - a_{n,j}
 \boldsymbol{w}_n , \{n+1 \le j \le M\} \cap \{j \notin {\mathcal S}\} .
 \end{array} \!\!\! \right\} \!\!\!
\end{equation}
 Then update $\zeta^{(n)}(k)=\zeta^{(n-1)}(k)-\big(w_n(k)\big)^2\big/
 \boldsymbol{w}_n^{\rm T}\boldsymbol{w}_n $ for $1\le k\le N$.
\\\hline
\end{tabular}
\end{center}
\vspace*{-3mm}
\end{table}

\begin{figure*}[ht]
\begin{center}
\includegraphics[width=0.33\linewidth]{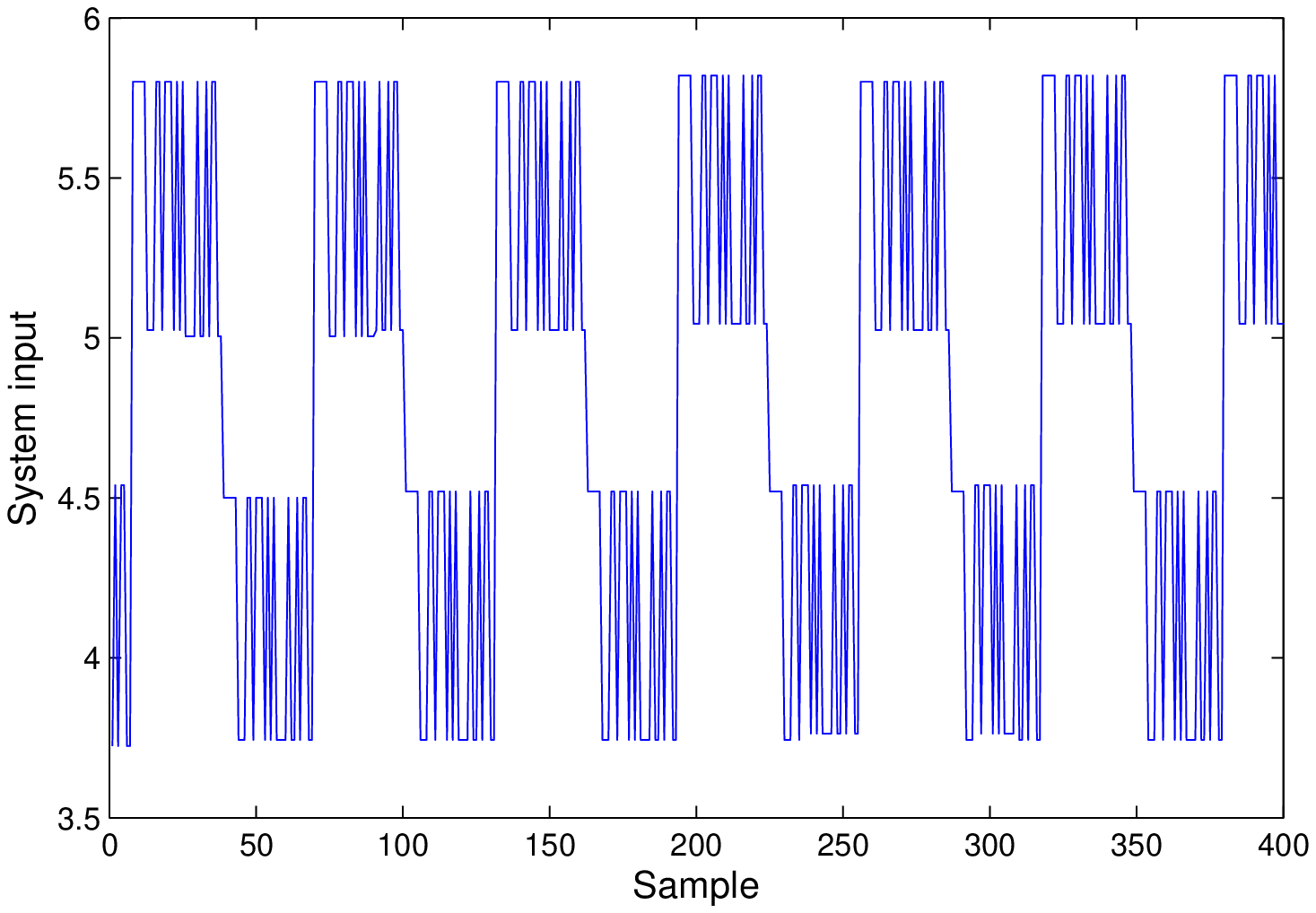}\includegraphics[width=0.33\linewidth]{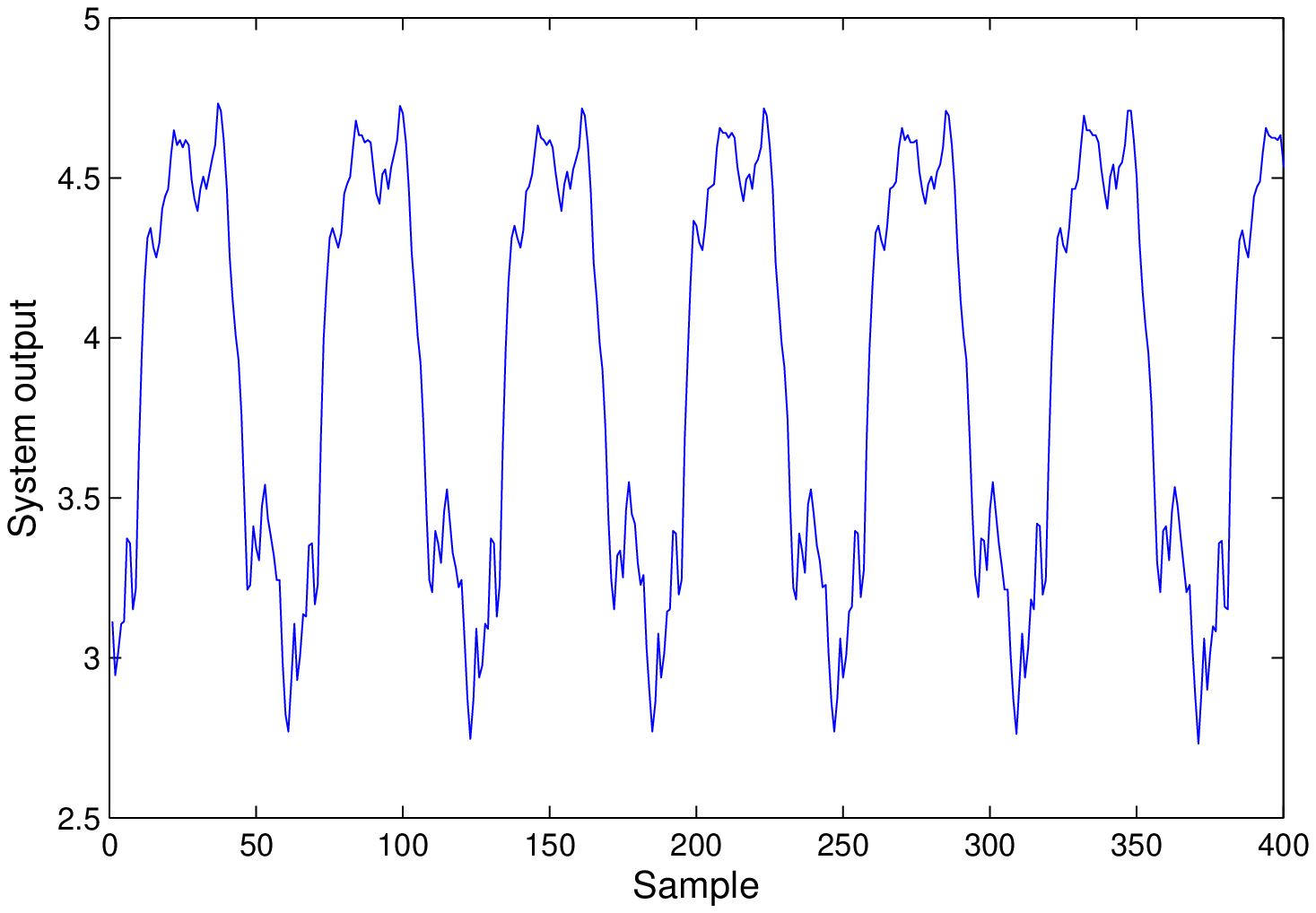}
\includegraphics[width=0.33\linewidth]{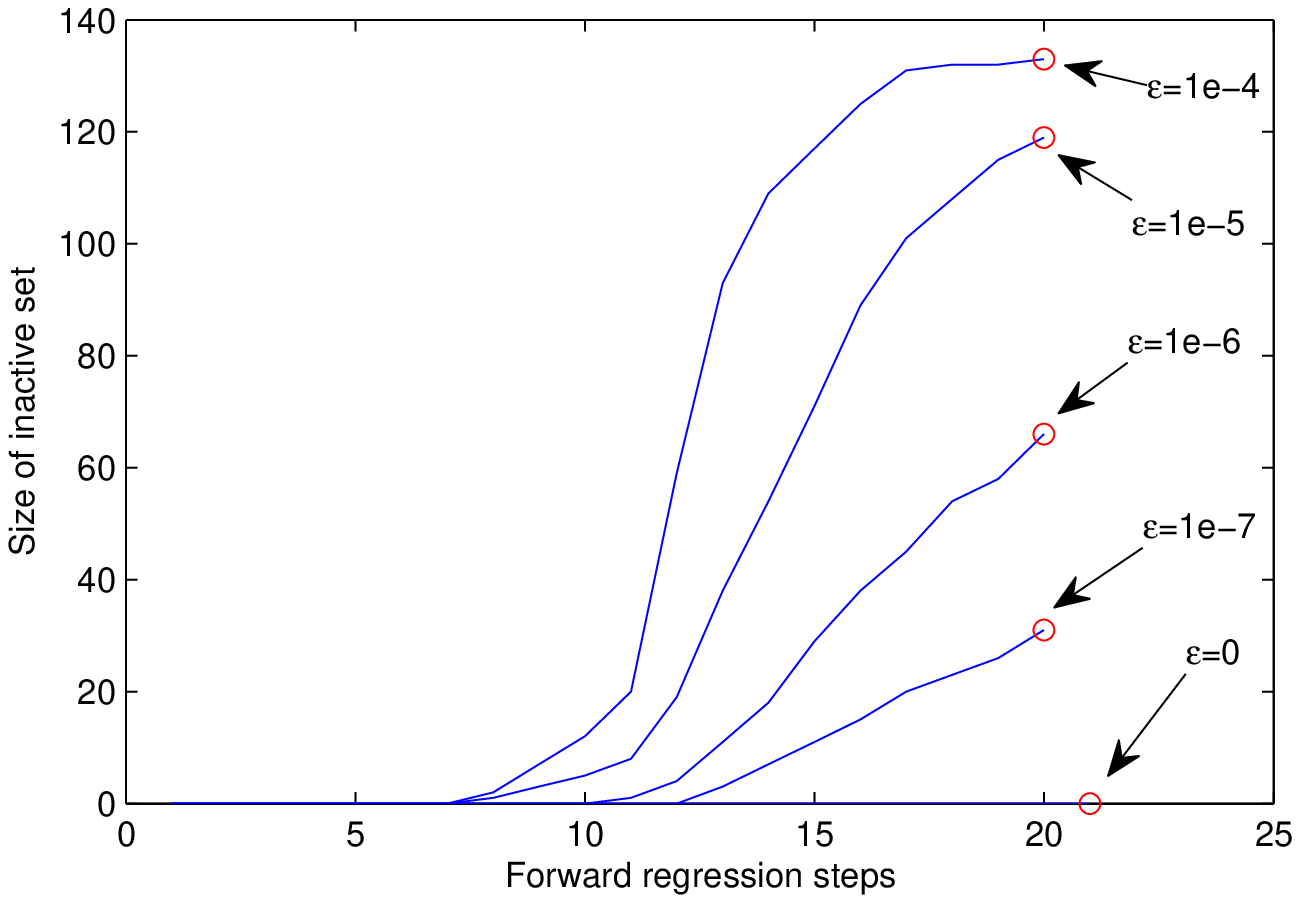}
\end{center}
 \vspace{-5mm}
\begin{center} {\small (a)}\hspace{55mm} {\small (b)}\hspace{55mm} {\small (c)}\end{center} 
 \vspace{-4mm}
\caption{Engine Data: (a)~the system input $u(k)$, (b)~the system output
 $y(k)$, and (c)~the evolution of the size of ${\mathcal S}$ with respect to the
 chosen $\varepsilon$.}
\label{fig:2}
 \vspace{-2mm}
\end{figure*}

 The initial conditions are as follows. Preset $\varepsilon >0$ as a very small value.
 Set $\boldsymbol{e}^{(0)}=\boldsymbol{y}$, $\zeta^{(0)}(k)=1$ for $1\le k\le N$, and
 ${\mathcal S}$ as the empty set $\emptyset$. The $n$th stage of the selection procedure
 is listed in Table~\ref{tab:1}. The OFR procedure is automatically terminated at the
 $(n_s+1)$th stage when the condition
\begin{align}\label{eq:51}
 J_{n_s +1} \ge J_{n_s}
\end{align}
 is detected, yielding a subset model with $n_s$ significant regressors. It is worth
 emphasizing that there always exists a model size $n_s$, and for $n \le n_s$, the
 LOOMSE $J_n$ decreases as $n$ increases, while the condition (\ref{eq:51}) holds
 \cite{Hong2003d,Chens2004a}.

 Note that the LOOMSE is used not only for deriving the closed form of the optimal
 regularization parameter estimate $\lambda_n^{\rm opt}$ but also for selecting the
 most significant model regressor. Specifically, a regressor is selected as the one 
 that produces the smallest LOOMSE value as well as offering the reduction in the
 LOOMSE. After the $n_s$ stage when there is no reduction in the LOOMSE criterion for
 a few consecutive OFR stages, the model construction procedure can be terminated.
 Thus, the $l^1$-POFR algorithm automatically constructs a sparse $n_s$-term model,
 where typically $n_s \ll M$.

 Also note that it is assumed that $\varepsilon$ should not be too small such that
 the LOOMSE estimation formula can be considered to be accurate. This means that if
 $\varepsilon$ is set too low, many insignificant candidate regressors will have
 inaccurate LOOMSE values for competition. However, we emphasize that these terms
 with inaccurate LOOMSE values will not be selected as the winner to enter the model.
 Hence in practice we only need to make sure that $\varepsilon$ is not too large,
 which would introduce unnecessary bias to the model parameter estimates. Clearly,
 a relatively larger $\varepsilon$ will save computational costs by 1)~resulting in
 a sparser model, and 2)~producing a larger sized inactive set during the OFR process.

 Finally, regarding the computational complexity of the $l^1$-POFR algorithm, if the
 unproductive regressors are not removed to the inactive set ${\mathcal S}$ during
 the OFR procedure, it is well known that the computational cost is in the order of
 $\textsf{O}(N)$ for evaluating each candidate regressor \cite{Chens2004a}. The total 
 computational cost then needs to be scaled by the number of evaluations in forward
 regression, which is $M(M-n_s)/2$. By removing unproductive regressors to ${\mathcal S}$
 during the OFR procedure, the computational cost can obviously be reduced significantly.
 It is not possible to exactly assess the computational cost saving due to removing the
 unproductive regressors, as this is problem dependent.

\begin{table}[bp!]
\vspace*{-3mm}
\caption{Comparison of the modeling performance for Engine Data. The computational cost
 saving is based on the same size of model without removing unproductive regressors in
 the $l^1$-POFR.}
\label{tab:2}
\vspace{-4mm}
 \begin{center}
\begin{tabular}{|c|c|c|c|c|}
\hline
 Algorithm & MSE          & MSE      & Model & Cost \\
           & training set & test set & size  &  saving \\ \hline
 LROLS-LOO \cite{Chens2004a}         & $0.000453$ & $0.000490$ & $22$  & NA \\ \hline
 $\varepsilon$-SVM ($\tau =3$)       & $0.000502$ & $0.000482$ & $208$ & NA \\
 $\varepsilon$-SVM ($\tau =2.5$)     & $0.000480$ & $0.000475$ & $208$ & NA \\
 $\varepsilon$-SVM ($\tau =2$)       & $0.000461$ & $0.000486$ & $208$ & NA \\
 $\varepsilon$-SVM ($\tau =1.5$)     & $0.000415$ & $0.000579$ & $208$ & NA \\
 $\varepsilon$-SVM ($\tau =1$)       & $0.000370$ & $0.000794$ & $208$ & NA \\ \hline
 LASSO ($\tau =1.5$)                 & $0.000923$ & $0.001010$ & $70$  & NA \\
 LASSO ($\tau =1$)                   & $0.000708$ & $0.000748$ & $44$  & NA \\ 
 LASSO ($\tau =0.5$)                 & $0.000706$ & $0.000842$ & $54$  & NA \\
 LASSO ($\tau =0.2$)                 & $0.000565$ & $0.000800$ & $81$  & NA \\
 LASSO ($\tau =0.1$)                 & $0.000644$ & $0.001907$ & $76$  & NA \\ \hline
 $l^1$-POFR ($\varepsilon =10^{-4}$) & $0.000498$ & $0.000502$ & $20$  & 27\% \\
 $l^1$-POFR ($\varepsilon =10^{-5}$) & $0.000492$ & $0.000480$ & $20$  & 18\% \\
 $l^1$-POFR ($\varepsilon =10^{-6}$) & $0.000484$ & $0.000485$ & $20$  & 8\%  \\
 $l^1$-POFR ($\varepsilon =10^{-7}$) & $0.000481$ & $0.000476$ & $20$  & 3\%  \\
 $l^1$-POFR ($\varepsilon =0$)       & $0.000452$ & $0.000472$ & $21$  & 0\%  \\ \hline
\end{tabular}
\end{center}
\vspace{-1mm}
\end{table}

\section{Simulation Study}\label{S5}

 \emph{Example 1}: This Engine Data set \cite{Billings1989b} contains the 410 data samples
 of the fuel rack position (the input $u(k)$) and the engine speed (the output $y(k)$),
 collected from a Leyland TL11 turbocharged, direct injection diesel engine which was
 operated at a low engine speed. The 410 input and output data points of the engine
 data set are plotted in Fig.~\ref{fig:2}~(a) and (b), respectively. The first $210$
 data samples were used in training and the last $200$ data samples for model testing.
 The previous study has shown that the data set can be modeled adequately using the
 system input vector $\boldsymbol{x}(k)=\big[y(k-1) ~ u(k-1) ~ u(k-2)]^{\rm T}$, and
 the best Gaussian RBF model was provided by the $l^2$-norm local regularization
 assisted OLS (LROLS) algorithm based on the LOOMSE (LROLS-LOO) \cite{Chens2004a} which
 was quoted in Table~\ref{tab:2} for comparison. The $\varepsilon$-SVM algorithm
 \cite{Gunn1998a} and the LASSO were also experimented based on the Gaussian kernel
 with a common variance $\tau^2$. For the $\varepsilon$-SVM, the Matlab function
 \emph{quadprog.m} was used with the algorithm option set as `interior-point-convex'. The
 tuning parameters in the $\varepsilon$-SVM algorithm, such as soft margin parameter
 $C$ \cite{Gunn1998a}, were set empirically so that the best possible result was obtained
 after several trials. For the LASSO, the Matlab function \emph{lasso.m} was used with
 10-fold CV being used to select the associated regularization parameter. For both the
 $\varepsilon$-SVM and LASSO, we list the results obtained for a range of kernel width
 $\tau$ values in Table~\ref{tab:2}, for comparison.

 Similar to the LROLS-LOO algorithm \cite{Chens2004a}, we also used the Gaussian RBF kernel
 (\ref{eq:3}) for the proposed $l^1$-POFR algorithm with an empirically set $\tau =2.5$ and
 the RBF centers $\boldsymbol{ c}_i$ were formed using all the training data samples. With a
 preset value of $\varepsilon$, a sparse model of size $n_s$ was automatically selected
 when the condition (\ref{eq:51}) was met. Fig.~\ref{fig:2}~(c) illustrates the evolution
 of the size of ${\mathcal S}$ with respect to a range of the preset $\varepsilon$ values.
 The test MSE values produced by the sparse models and the sizes of the models associated
 with the same range of $\varepsilon$ values are recorded in Table~\ref{tab:2}, which show
 that the excellent model generalization capability of all the models generated by the
 proposed algorithm. Moreover, the $l^1$-POFR algorithm produces the sparsest model.

\begin{table}[bp!]
\vspace*{-3mm}
\caption{Comparison of the modeling performance for Boston House Data. The results were
 averaged over 100 realizations and given as $\mbox{mean}\pm\mbox{standard deviation}$.}
\label{tab:3}
\vspace{-4mm}
 \begin{center}
\begin{tabular}{|l|c|c|c|}
\hline
 Algorithm & MSE          & MSE     & Model  \\
           & training set & test set & size  \\ \hline
 $\varepsilon$-SVM \cite{Gunn1998a}  & $6.80\pm 0.44$  & $23.18\pm 9.05$ & $243\pm 5.3$   \\ \hline
 LROLS-LOO \cite{Chens2004a}         & $12.97\pm 2.67$ & $17.42\pm 4.67$ & $58.6\pm 11.3$ \\ \hline
 NonOFR-LOO \cite{Chens2009a}        & $10.10\pm 3.40$ & $14.07\pm 3.62$ & $34.6\pm 8.4$  \\ \hline
 LASSO ($\tau =2$)                   & $8.52\pm 3.57$  & $14.37\pm 8.15$ & $76.8\pm 39.7$ \\ 
 LASSO ($\tau =3$)                   & $8.55\pm 1.07$  & $13.31\pm 6.65$ & $68.6\pm 29.3$ \\ 
 LASSO ($\tau =5$)                   & $10.45\pm 1.07$ & $15.05\pm 8.37$ & $85.9\pm 19.7$ \\ 
 LASSO ($\tau =10$)                  & $16.42\pm 1.78$ & $19.39\pm 8.31$ & $29.9\pm 21.3$ \\ \hline
 $l^1$-POFR ($\varepsilon =0.01$)    & $9.99\pm 1.37$  & $14.47\pm 7.47$ & $30.5\pm 5.3$ \\
 $l^1$-POFR ($\varepsilon =0.001$)   & $9.24\pm 1.57$  & $14.10\pm 7.02$ & $34.9\pm 7.8$ \\
 $l^1$-POFR ($\varepsilon =0.0001$)  & $9.07\pm 1.64$  & $14.02\pm 6.85$ & $36.6\pm 9.3$ \\
 $l^1$-POFR ($\varepsilon =0.00001$) & $9.08\pm 1.64$  & $13.95\pm 6.76$ & $36.5\pm 9.3$ \\ \hline
\end{tabular}
\end{center}
\vspace*{-1mm}
\end{table}

\emph{Example 2}: This regression benchmark data set, Boston Housing Data, is available
 at the UCI repository \cite{Frank+Asuncion:2010}. The data set comprises 506 data
 points with 14 variables. The previous study \cite{Chens2009a} performed the task
 of predicting the median house value from the remaining 13 attributes using the
 $\varepsilon$-SVM \cite{Gunn1998a}, the LROLS-LOO \cite{Chens2004a} and the nonlinear
 OFR based on the LOOMSE (NonOFR-LOO) \cite{Chens2009a}. The NonOFR-LOO algorithm
 \cite{Chens2009a} constructs a \emph{nonlinear} RBF model in the OFR procedure, where
 each stage of the OFR determines one RBF node's center vector and diagonal covariance
 matrix by minimizing the LOOMSE. In the experiment study presented in \cite{Chens2009a},
 456 data points were randomly selected from the data set for training and the remaining
 50 data points were used to form the test set. Average results were given over 100
 realizations. For each realization, 13 input attributes were normalized so that each
 attribute had zero mean and standard deviation of one. We also experimented with the
 LASSO supplied by Matlab \emph{lasso.m} with option set as 10-fold CV to select the
 associated regularization parameter. For the LASSO, a common kernel width $\tau$ was
 set for constructing the kernel model from the 456 candidate regressors of each
 realization, and a range of $\tau$ values were experimented.

 For the $l^1$-POFR, $\tau =15$ was empirically set for constructing 456 candidate
 Gaussian RBF regressors of each realization. We experimented a range of the preset
 $\varepsilon$ values for the $l^1$-POFR algorithm, and the results obtained are as
 summarized in Table~\ref{tab:3}, in comparison with the results obtained by the
 $\varepsilon$-SVM and the LASSO, as well as the LROLS-LOO and NonOFR-LOO, which are
 quoted from the study \cite{Chens2009a}.

\section{Conclusions}\label{S6}

 We have developed an efficient data model algorithm, referred to as the $l^1$-norm
 penalized orthogonal forward regression ($l^1$-POFR), for linear-in-the-parameters
 nonlinear models based on a new $l^1$-norm penalized cost function defined in the
 constructed orthogonal modeling space. The LOOMSE is used for simultaneous model
 term selection and regularization parameter estimation in a highly efficient OFR
 procedure. Additionally, we have proposed a lower bound of the regularisation
 parameters for robust LOOMSE estimation as well as detecting and removing 
 insignificant regressors to an inactive set along the OFR process, further enhancing
 the efficiency of the OFR procedure. Numerical studies have been utilized to
 demonstrate the effectiveness of this new $l^1$-POFR approach.

%\bibliographystyle{ieee}        % Include this if you use bibtex
%\bibliography{nev_ref}           % and a bib file to produce the

\end{document}